\documentclass[10pt,twocolumn,letterpaper]{article}

\usepackage{iccv}
\usepackage{times}
\usepackage{epsfig}
\usepackage{graphicx}
\usepackage{amsmath}
\usepackage{amssymb}

\usepackage{subfigure}
\usepackage{multirow}
\usepackage{url}
\usepackage{booktabs}
\usepackage{arydshln}

\usepackage[pagebackref=true,breaklinks=true,letterpaper=true,colorlinks,bookmarks=false]{hyperref}

\iccvfinalcopy % *** Uncomment this line for the final submission

\def\iccvPaperID{2247} % *** Enter the ICCV Paper ID here

\begin{document}

%%%%%%%%% TITLE
\title{Denoising Diffusion Autoencoders are Unified Self-supervised Learners}

\author{Weilai Xiang$^{1,2}$ \quad Hongyu Yang$^{1,3}$\thanks{Corresponding author.} \quad Di Huang$^{2}$ \quad Yunhong Wang$^{1,2}$\\
$^{1}${\normalsize State Key Laboratory of Virtual Reality Technology and Systems, Beihang University, Beijing, China}\\
$^{2}${\normalsize School of Computer Science and Engineering, Beihang University, Beijing, China}\\
$^{3}${\normalsize Institute of Artificial Intelligence, Beihang University, Beijing, China}\\
{\tt\small \{xiangweilai, hongyuyang, dhuang, yhwang\}@buaa.edu.cn}
}

\maketitle
\ificcvfinal\thispagestyle{empty}\fi

\begin{abstract}
Inspired by recent advances in diffusion models, which are reminiscent of denoising autoencoders, we investigate whether they can acquire discriminative representations for classification via generative pre-training.
This paper shows that the networks in diffusion models, namely denoising diffusion autoencoders (DDAE), are \textbf{unified} self-supervised learners:
by pre-training on unconditional image generation, DDAE has already learned strongly linear-separable representations within its intermediate layers without auxiliary encoders, thus making diffusion pre-training emerge as a general approach for generative-and-discriminative dual learning.
To validate this, we conduct linear probe and fine-tuning evaluations. Our diffusion-based approach achieves 95.9\% and 50.0\% linear evaluation accuracies on CIFAR-10 and Tiny-ImageNet, respectively, and is comparable to contrastive learning and masked autoencoders for the first time.
Transfer learning from ImageNet also confirms the suitability of DDAE for Vision Transformers, suggesting the potential to scale DDAEs as unified foundation models. Code is available at \href{https://github.com/FutureXiang/ddae}{github.com/FutureXiang/ddae}.
\end{abstract}

\section{Introduction}
\label{sec:intro}
Understanding data with limited human supervision is a crucial challenge in machine learning. To cope with massive amounts of data with scarce annotations, deep learning paradigms are shifting from supervised to self-supervised pre-training. 
Regarding natural language processing (NLP), self-supervised models such as BERT \cite{bert}, GPTs \cite{gpt, gpt2, gpt3} and T5 \cite{t5} have achieved outstanding performance across diverse tasks, and large language models like ChatGPT \cite{gptHF} are showing a profound impact beyond the machine learning community.
Among these, BERT uses masked language modeling (MLM) as a pretext task to train encoders while which cannot generate full text samples. In contrast, GPTs and T5 have shown capabilities in generating long paragraphs autoregressively (AR). Moreover, they prove that decoder-only or encoder-decoder models can acquire deep language understandings via generative pre-training, without the need of training an encoder intentionally.
With the rise of AI-Generated Content (AIGC), GPTs and T5 have been garnering more attention compared to pure encoders, which \textbf{unify} the generative (\eg translation, summarization) and discriminative (\eg classification) tasks \cite{gpt2, t5}.

\begin{figure}[t]
\centering
    \subfigure[Denoising networks in pixel-space and latent-space diffusion models.]{
    \includegraphics[width=\linewidth]{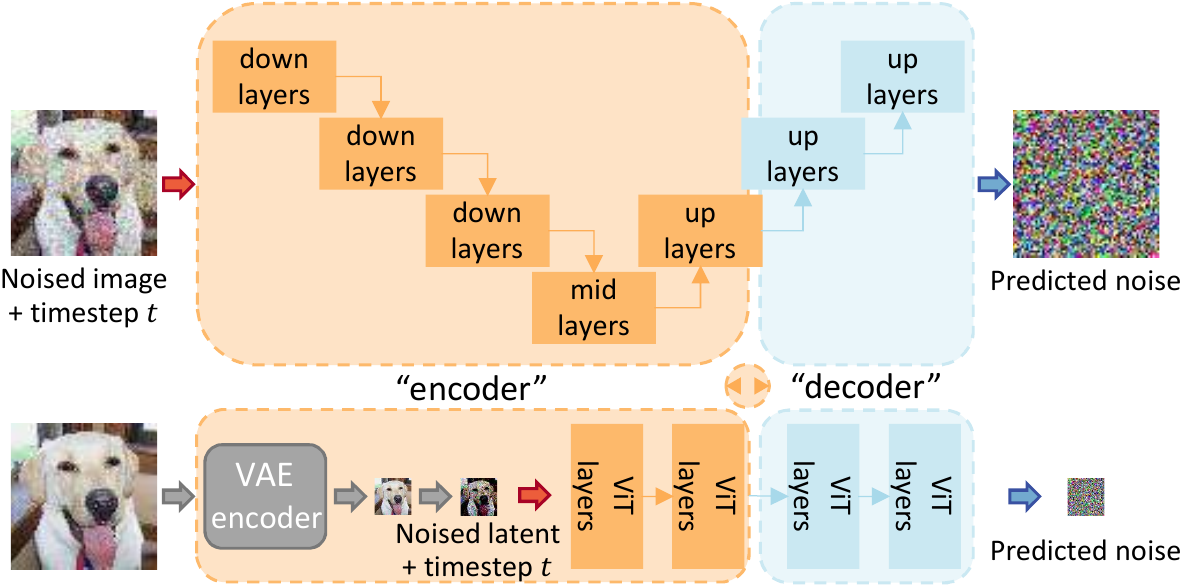}
    \label{fig:DDAE_1}
    }
    \subfigure[Evaluating DDAEs as self-supervised representation learners.]{
    \includegraphics[width=\linewidth]{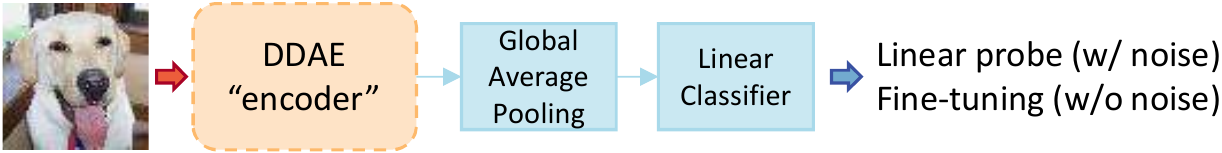}
    \label{fig:DDAE_2}
    }
    \caption{\textbf{Denoising Diffusion Autoencoders (DDAE).}
    \textit{Top:} Diffusion networks are essentially equivalent to level-conditional denoising autoencoders (DAE). The networks are named as DDAEs due to this similarity.
    \textit{Bottom:} By linear probe evaluations, we confirm that DDAE can produce strong representations at some intermediate layers. Truncating and fine-tuning DDAE as vision encoders further leads to superior image classification performance.
    }
\end{figure}

In computer vision, self-supervised learners have not yet achieved similar feats as GPTs in bridging the gap between generation and recognition. While generative adversarial networks (GAN) \cite{gan, stylegan2, biggan} and autoregressive Transformers \cite{transformer, vqgan, dalle} can synthesize high-fidelity images, they do not offer significant benefits for discriminative tasks.
For recognition, contrastive methods \cite{simclr, simclr2, byol, dino} build features through pretext tasks on augmented images. Masked autoencoders \cite{mae, beit, simmim} introduce BERT-like masked image modeling (MIM) pre-training for Vision Transformers \cite{vit}, but they seem not natural and practical for convolutional networks.
Note that although MIM-based methods can recover masked image tokens, they are problematic in synthesizing full images, mainly because the complete data distribution is not directly modeled.
These methods are not referred to as unified generative-and-discriminative models. Instead, we denote them as ``semi-'' generative for their partial similarities to full generative models (Table~\ref{tab:ssl}).

Theoretically, it is more practical to extend generative models for discriminative purposes and gain the benefit of both ways.
Recently, we have witnessed the flourish of AI-generated visual content due to the emergence of diffusion models \cite{ddpm, sde}, with state-of-the-art results reported in image synthesis \cite{adm, edm, dit}, image editing \cite{prompt2prompt} and text-to-image synthesis \cite{dalle2, ldm, imagen}.
Considering such capability, versatility, and scalability in generative modeling, we ask: \textit{whether diffusion models can replicate the success of GPTs and T5, in becoming unified generative-and-discriminative learners?} We regard them as very promising alternatives in the visual domain, based on the following observations:

\textbf{(\romannumeral1)} It has been demonstrated that discriminative image or text encodings can be learned through end-to-end generative pre-training \cite{igpt, t5-enc, t5-sts}, \ie ``analysis-by-synthesis''.
Intuitively, the full generation task should contain, and be more challenging than the semi-generative masked modeling, suggesting that image (or language) generation is compatible with visual (or language) understanding, but not vice versa.
The \textbf{generative pre-training} paradigm supports diffusion as a meaningful discriminative learning method.

\textbf{(\romannumeral2)} Diffusion networks are trained as multi-level denoising autoencoders (DAE, Figure~\ref{fig:DDAE_1}). The idea of \textbf{denoising autoencoding} has been widely applied to discriminative visual representation learning \cite{dae1, dae2, seg-decoder-denoise}.
More recently, masked autoencoders (MAE) \cite{mae} have further highlighted the effectiveness of denoising pre-training, which can also be inherited by diffusion networks --- likewise, recovering images with large, multi-scale noise is nontrivial and may also require a high-level understanding of visual concepts.

\textbf{(\romannumeral3)} The benefits of diffusion-based representation learning are evidenced.
DDPM-seg \cite{seg-diffusion} confirms that diffusion can capture pixel-wise semantic information, indicating the feasibility.
Besides, previous attempts in GANs and Transformers, \ie BigBiGAN \cite{bigbigan} and iGPT \cite{igpt}, find that better image generation capability can translate to improved feature quality, suggesting that diffusion is even more capable of representation learning as the state-of-the-art generative model.
Diffusion-based representation learners can also be facilitated by large AIGC projects if the learned knowledge can be conveniently transferred from pre-trained models.

\begin{table}[]
\resizebox{\linewidth}{!}{
    \begin{tabular}{llll}
    \hline
    \multirow{2}{*}{\textbf{Model}} & \multirow{2}{*}{\begin{tabular}[c]{@{}l@{}}\textbf{Pre-training} \\ \textbf{target and method}\end{tabular}} & \multirow{2}{*}{\begin{tabular}[c]{@{}l@{}}\textbf{(\romannumeral1) Generative} \\ \textbf{pre-training}\end{tabular}} & \multirow{2}{*}{\begin{tabular}[c]{@{}l@{}}\textbf{(\romannumeral2) Denoising}\\ \textbf{autoencoding}\end{tabular}} \\
     & & & \\ \hline
    \multicolumn{4}{l}{\textit{Natural   Language Processing}}      \\
    BERT \cite{bert}       & Encoder-only,  MLM & Semi-  & Masked   \\
    GPT \cite{gpt}         & Decoder-only,  AR  & Full   & --       \\
    T5 \cite{t5}           & Enc-dec, MLM+AR    & Full   & Masked   \\ \hline
    \multicolumn{4}{l}{\textit{Computer   Vision}}                  \\
    MAE \cite{mae}         & Encoder-only,  MIM & Semi-  & Masked   \\
    iGPT \cite{igpt}       & Decoder-only,  AR  & Full   & --       \\
    \multirow{2}{*}{\textbf{DDAE (ours)}} & \multirow{2}{*}{Enc-dec, Diffusion} & \multirow{2}{*}{Full} & \multirow{2}{*}{\begin{tabular}[c]{@{}l@{}}Multi-level \\ Gaussian\end{tabular}} \\
     & & & \\ \hline \\
    \end{tabular}
}
\caption{\label{tab:ssl} A summary of self-supervised learners. DDAE takes full advantage of generative pre-training and denoising autoencoding.}
\end{table}

Driven by this analysis, we investigate whether diffusion models, which incorporate the best practices of generative pre-training and denoising autoencoding (as summarized in Table~\ref{tab:ssl}), can learn effective representations for image classification.
Our approach is straightforward: we evaluate diffusion pre-trained networks, namely denoising diffusion autoencoders (DDAE), as feature extractors by measuring the linear probe and fine-tuning accuracies of intermediate activations (Figure~\ref{fig:DDAE_2}).
For linear probing, we pass noised images with specific scales (or timesteps) to DDAE and examine the activations at different layers. For fine-tuning, we truncate DDAE at the best representation layer as an image encoder and fine-tune it without additional noising.

We confirm that via end-to-end diffusion pre-training, DDAEs do learn strongly linear-separable features, which lie in the middle of up-sampling and can be extracted when images are perturbed with noises.
Moreover, we validate the correlation between generative and discriminative performance of DDAEs through ablation studies on noise configurations, training steps, and the mathematical model.
Evaluations on CIFAR-10 \cite{cifar10} and Tiny-ImageNet \cite{tiny} show that the diffusion-based approach is comparable to supervised WideResNet \cite{wrn}, contrastive SimCLRs \cite{simclr, simclr2} and MAE \cite{mae} for the first time. The transfer ability has also been verified on ImageNet \cite{in1k} pre-trained models, including the ones constructed by pixel-space UNets and latent-space Vision Transformers such as DiT \cite{dit}.

Our study highlights the underlying nature of diffusion models as unified vision foundation models. The revealed duality of DDAEs as state-of-the-art generative models and competitive recognition models may inspire improvements to vision pre-training and applications in both domains.
With the insightful elucidation and observations presented in this paper, it is highly likely to transfer powerful discriminative knowledge from large-scale pre-trained AIGC models like Stable Diffusion \cite{ldm} in the near future.

\section{Related work}
\textbf{Diffusion models} are becoming the most popular generative paradigm due to their high-fidelity performance and the ability to synthesize complex visual concepts \cite{ddpm, sde, edm, ldm, dalle2, imagen}, without unstable adversarial training, mode collapse issues or architecture constraints.
With improvements to computational efficiency \cite{ldm}, training \cite{edm}, sampling \cite{ddim, iddpm, edm} and guidance \cite{adm, cfg}, diffusion models become the state-of-the-art on unconditional CIFAR-10 \cite{cifar10}, class-conditional ImageNet \cite{in1k}, and text-to-image on MS-COCO \cite{coco}. AIGC projects such as Stable Diffusion \cite{ldm} and ControlNet \cite{ctrlnet} have achieved broad social impacts.
Recent work on diffusion with ViTs \cite{vit, uvit, dit} further explores diffusion models with scalable backbones.

\textbf{Representation learning with generative models} is a long-standing idea since they model the data distribution in an unsupervised manner. While VAEs \cite{vae, vqvae} can learn meaningful representations, they have proven more useful in generation than recognition.
BigBiGAN \cite{bigbigan} learns discriminative features from large-scale GANs \cite{biggan} with jointly trained encoders \cite{bigan1, bigan2}. However, the feature quality may be compromised since GANs naturally capture less data diversity.
iGPT \cite{igpt} models next pixel prediction with Transformers and achieves competitive results with contrastive methods, but the lack of inductive bias for images makes its downstream applications restricted and inefficient.

\textbf{Representation learning with diffusion models.} A number of studies \cite{diff-ae, diff-ae-pretrained} introduce auxiliary encoders to extract representations following GAN Inversion \cite{ganinv}, but they focus on attribute manipulation rather than recognition.
Other methods learn linear-separable features with modified diffusion frameworks \cite{DRL-VDRL, DRL-VDRL-time}, while they underperform contrastive baselines by large margins.
Training diffusion models with classification objectives has been explored \cite{hybrid}, but it fails to rival pure recognition models and hurts generative performance heavily.
Diffusion-based conditional likelihood estimation \cite{score-class} is also straightforward, but its accuracy on CIFAR-10 is still below ResNet.
In contrast, our approach is comparable to typical self-supervised and supervised models without modifying diffusion frameworks.
Our study is partially inspired by DDPM-seg, which studies DDPM \cite{ddpm} on single super-class datasets for segmentation. However, we are the first to explore various diffusion models for classification on complex multi-class datasets.

\section{Approach}
\subsection{Background: DDAEs as generative models}
\label{sec:background}
Diffusion models \cite{ddpm,sde,edm} define a series of data corruptions which apply Gaussian noise to data $x_0$. Given timestep $t=1,...,T$ which indicates noise levels, the corruption is defined as $q(x_t | x_0) = \mathcal{N}(\alpha_t x_0, \sigma_t^2 \mathbf{I})$, where $\alpha_t$ and $\sigma_t$ are hyper-parameters controlling the signal-to-noise ratio.
When $T$ is large enough, data will be approximately corrupted to $x_T \sim \mathcal{N}(0, \mathbf{I})$.
With the reparameterization trick, a noised version of $x_0$ at an arbitrary level of $t$ can be obtained, by sampling $\epsilon \sim \mathcal{N}(0, \mathbf{I})$ and taking:
\begin{equation}
    x_t =  \alpha_t x_0 + \sigma_t \epsilon
    \label{eq:noising}
\end{equation}

Diffusion models aim to invert the corruption and reconstruct samples. Specifically, a random $x_T$ is drawn from $\mathcal{N}(0, \mathbf{I})$, and the model samples $x_{t-1} \sim q(x_{t-1} | x_t)$ iteratively until getting $x_0$.
Diffusion models employ trainable networks to approximate the transitions with $p_\theta(x_{t-1} | x_t) = \mathcal{N}(\mu_\theta(x_t, t), \Sigma_t^2 \mathbf{I})$, whose mean $\mu_\theta(x_t, t)$ is predicted by networks and variance $\Sigma_t^2$ is a constant. By simplifying the maximize likelihood objective, networks are equivalently trained with the \textbf{denoising autoencoder objective} \cite{edm}:
\begin{equation}
    \mathcal{L}_{denoise} = \|D_\theta(x_t, t) - x_0\|^2
    \label{eq:denoise}
\end{equation}
where $D_\theta(x_t, t)$ is a function of $x_t$ and $\mu_\theta(x_t, t)$. Therefore, a denoising network trained by diffusion modeling can be seen as a multi-level and level-conditional version of DAEs, specifically \textit{Denoising Diffusion Autoencoders (DDAE)}.

DDPM \cite{ddpm} proposes the ``Variance Preserving'' parameterization, where $\alpha_t = \sqrt{\Pi_{i=1}^{t}{(1-\beta_i)}}$ and $\alpha_t^2 + \sigma_t^2 = 1$. $\beta_{1...T}$ are determined by a linear schedule from $\beta_{min}$ to $\beta_{max}$. The network is trained to minimize the noise prediction error $\|\epsilon_\theta(x_t, t) - \epsilon\|^2$, which is a re-weighted version of Eq.~\ref{eq:denoise}, since the denoiser can be derived from $D_\theta(x_t, t) = \frac{x_t - \sigma_t \epsilon_\theta(x_t, t)}{\alpha_t}$. DDPM uses a UNet with 35.7M parameters, and achieves competitive performance on CIFAR-10.

EDM \cite{edm} proposes to use the ``Variance Exploding'' parameterization, where $\alpha_t = 1$ and $\sigma_t$ is sampled by an improved schedule. Based on score-based stochastic differential equations \cite{sde}, which extend the corruptions to infinite timesteps, EDM yields state-of-the-art results on CIFAR-10 using a larger DDPM++ network \cite{sde} with 56M parameters.

Apart from the mathematical formulation improvements, some studies explore efficient and scalable architectures for DDAEs. In particular, Latent Diffusion Models (LDM) \cite{ldm} perform diffusion modeling within the compressed VAE latent space and outperform pixel-space models on high-resolution text-to-image synthesis. DiT \cite{dit} explores scalable backbones under the LDM framework, which replaces UNets with Vision Transformers and achieves state-of-the-art results on class-conditional ImageNet generation.

In this paper, we consider (\romannumeral1) the original DDPM UNet, (\romannumeral2) DDPM++ trained by EDM, and (\romannumeral3) latent-space DiT as representative DDAE implementations. DDPM(++) uses UNets with timestep embeddings to parameterize $\epsilon_\theta(x_t, t)$ or $D_\theta(x_t, t)$. DiT uses latent-space ViTs with timestep and label embeddings to learn $\epsilon_\theta(x_t, y, t)$ for conditional generation, and jointly learns an unconditional model $\epsilon_\theta(x_t, \O, t)$ to achieve classifier-free guidance \cite{cfg} sampling.

\subsection{Evaluating DDAEs as discriminative learners}
\label{sec:extracting}
Extracting meaningful and discriminative representations from DDAEs is not trivial. Although deterministic inference methods like DDIM \cite{ddim} are able to derive uniquely identifiable encodings, the contained information is not compact enough for classification. There also exists a trend to employ additional encoders to learn representations for attribute manipulation \cite{diff-ae, diff-ae-pretrained} or classification \cite{DRL-VDRL, DRL-VDRL-time}.
In contrast, inspired by iGPT \cite{igpt} and DDPM-seg \cite{seg-diffusion} which evaluate learned features in autoregressive Transformers and diffusion UNets, we propose to \textit{directly take the intermediate activations in pre-trained DDAEs}. This approach does not require modification to common diffusion frameworks and is compatible with all existing models.

Drawing from the connections with denoising autoencoders, it is possible that DDAEs can produce linear-separable representations at some implicit encoder-decoder interfaces, resembling MAE.
Driven by this, we extend previous investigations on GPTs and DDPM \cite{igpt, seg-diffusion} to various network backbones (UNets and DiT) under different frameworks (DDPM and EDM).
Considering that UNet with skip connections has been the de-facto design, we avoid splitting the encoder-decoder explicitly to prevent diminishing the generation performance. However, the best layer to extract features remains unknown. Additionally, to prevent the gap between pre-training and deploying, images have to be noised by certain scales for linear evaluations.
Considering both the aforementioned facts, we investigate the relationship between feature qualities and layer-noise combinations through grid search, following DDPM-seg.

\begin{figure}[t]
\centering
    \subfigure{
    \includegraphics[width=\linewidth]{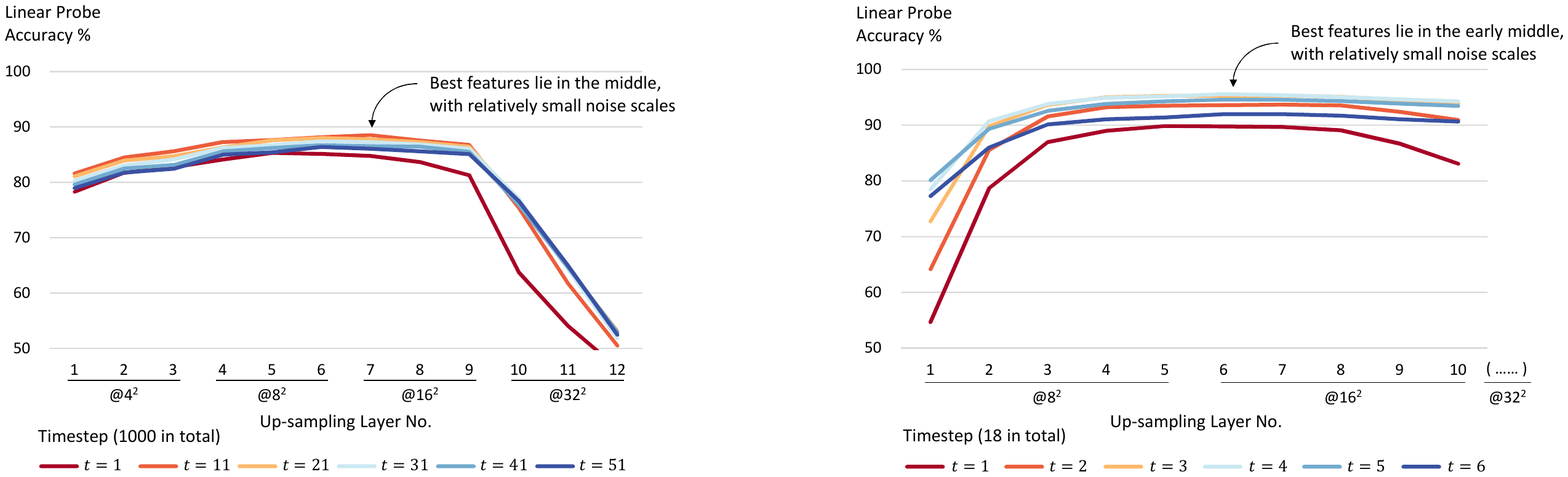}
    }
    \subfigure{
    \includegraphics[width=\linewidth]{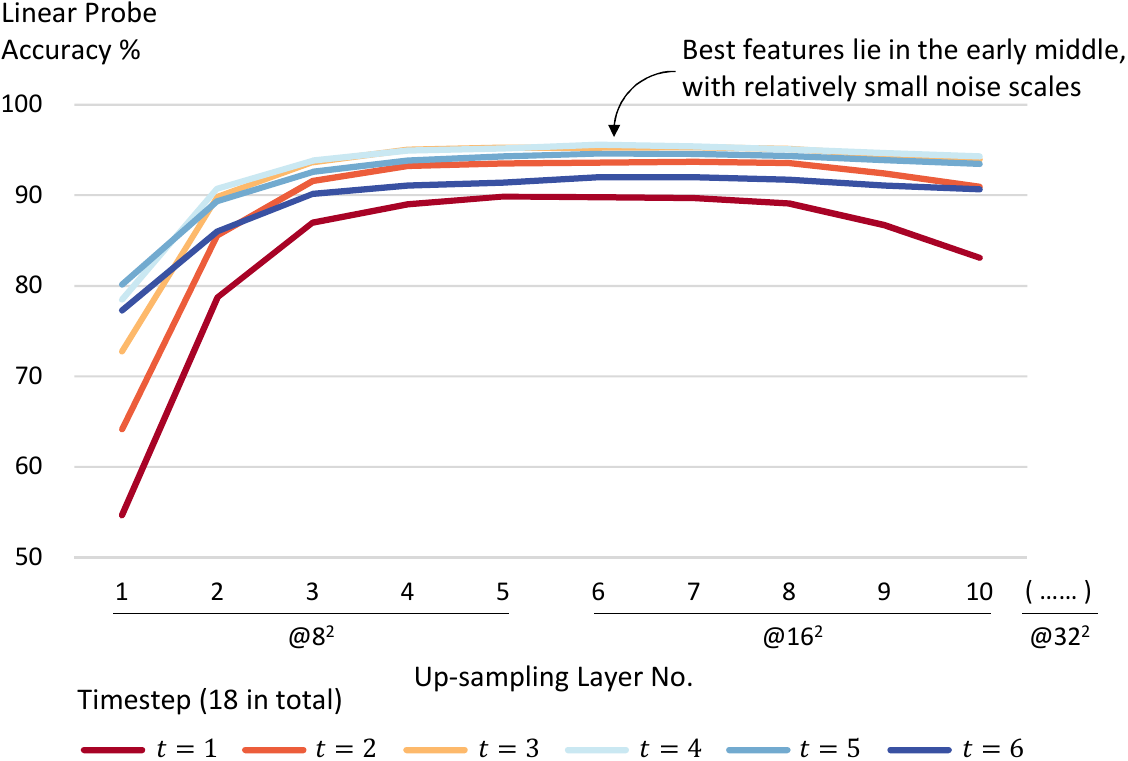}
    }
    \caption{\textbf{Feature quality depends on layer depths and noising scales.}
    Best features in DDAE UNets lie in the middle of the up-sampling stage on CIFAR-10, when images are perturbed with small noises. \textit{Top:} DDPM. \textit{Bottom:} DDPM++ trained by EDM.
    }
\label{fig:layer_time}
\end{figure}

To apply noise, we randomly sample $\epsilon$ and use Eq.~\ref{eq:noising} to obtain $x_t$, since no obvious differences can be observed between random and deterministic noising. Linear probe accuracies on the features after global average pooling are examined, as illustrated in Figure~\ref{fig:DDAE_2}.
Figure~\ref{fig:layer_time} shows that layer depths and noising scales affect feature quality jointly as a concave function, whose global maximum point can be found empirically. For the resolution of $32^2$ pixels, the best features lie in the middle of up-sampling, rather than at the lowest resolution as in common practices.
Furthermore, we find that perturbing images with relatively small noises improves the linear probe performance, especially on DDPM++ trained by EDM, which achieves {\boldmath$95\%+$} linear probe accuracy and surpasses classical AE or VAEs \cite{DRL-VDRL}. 

These properties have been verified across different datasets and models, but the optimal layer-noise combination may vary under different settings.
In the remainder of this paper, linear probe accuracies are reported as the highest found in grid search. For fine-tuning, clean images are passed to DDAE encoders, which are truncated at the optimal layers. The timestep inputs are also fixed to the optimal values. Note that this may not perform best for fine-tuning, since we find the fine-tuning accuracy can be further improved if more layers are used, and it is less sensitive to timestep inputs. However, we keep the layer-noise setup consistent with linear probing to reduce notation overhead.

\subsection{Label-free monitoring for layer selection}
\begin{figure*}[t]
\centering
    \subfigure{
    \includegraphics[height=4.3cm]{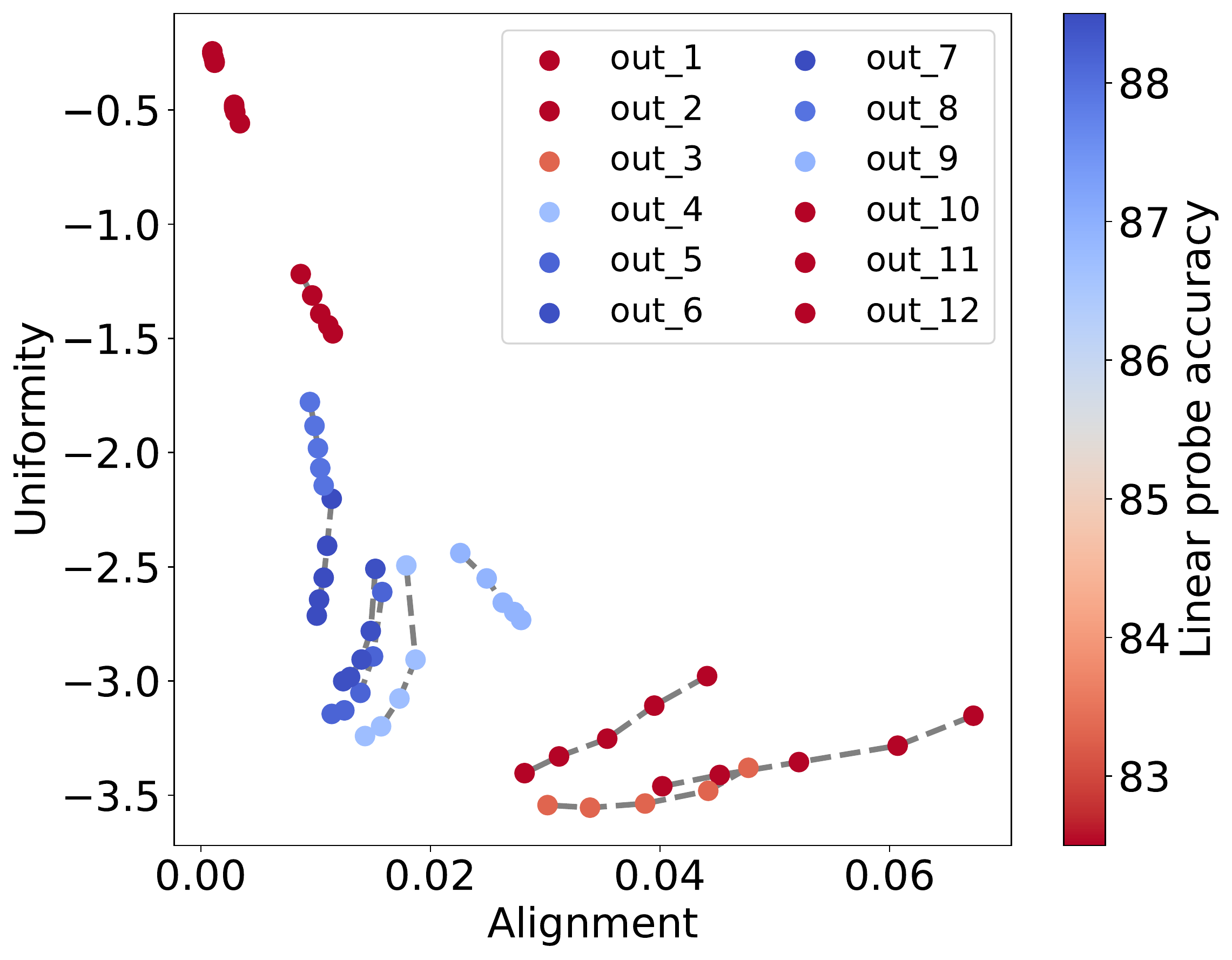}
    }
    \subfigure{
    \includegraphics[height=4.3cm]{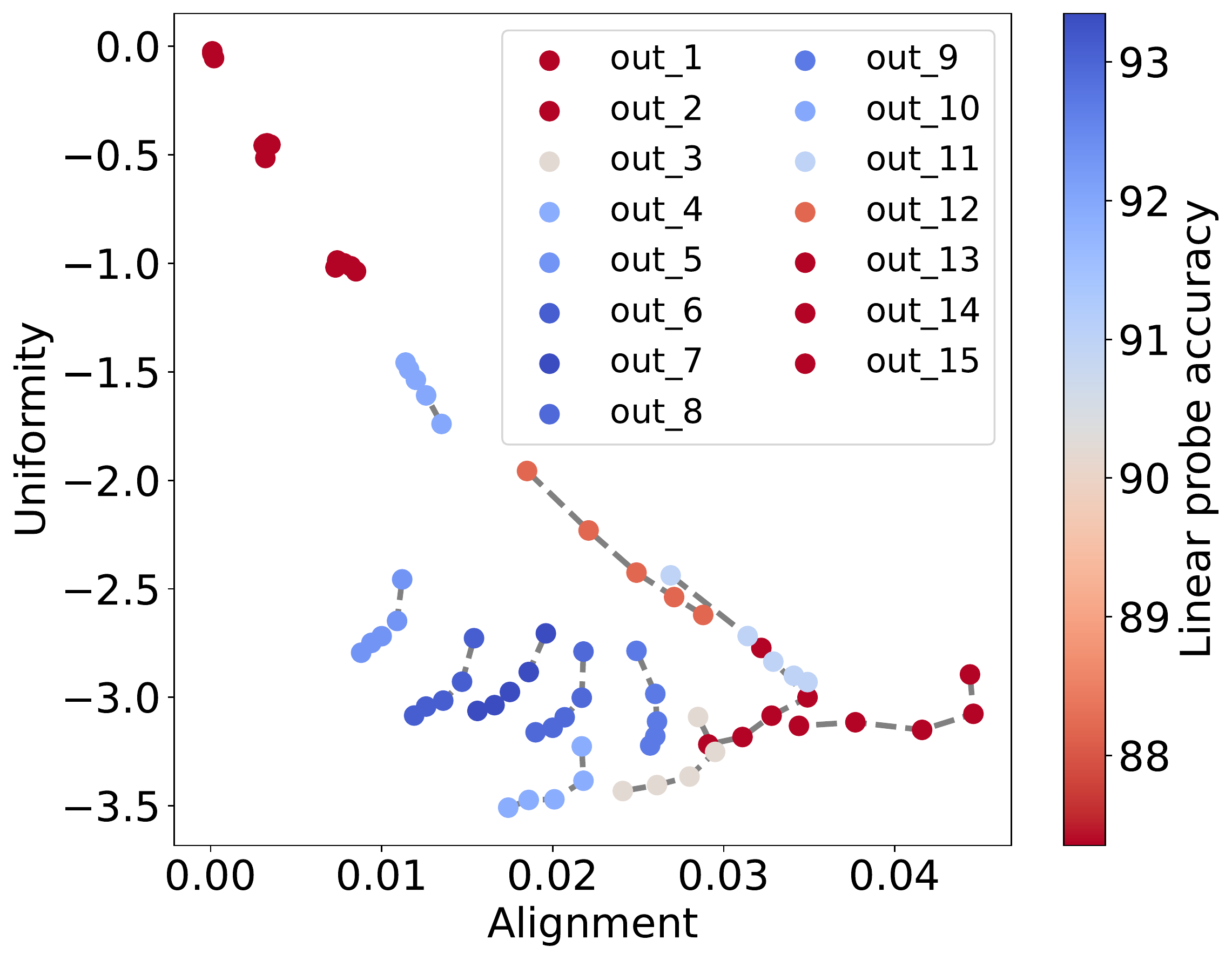}
    }
    \subfigure{
    \includegraphics[height=4.3cm]{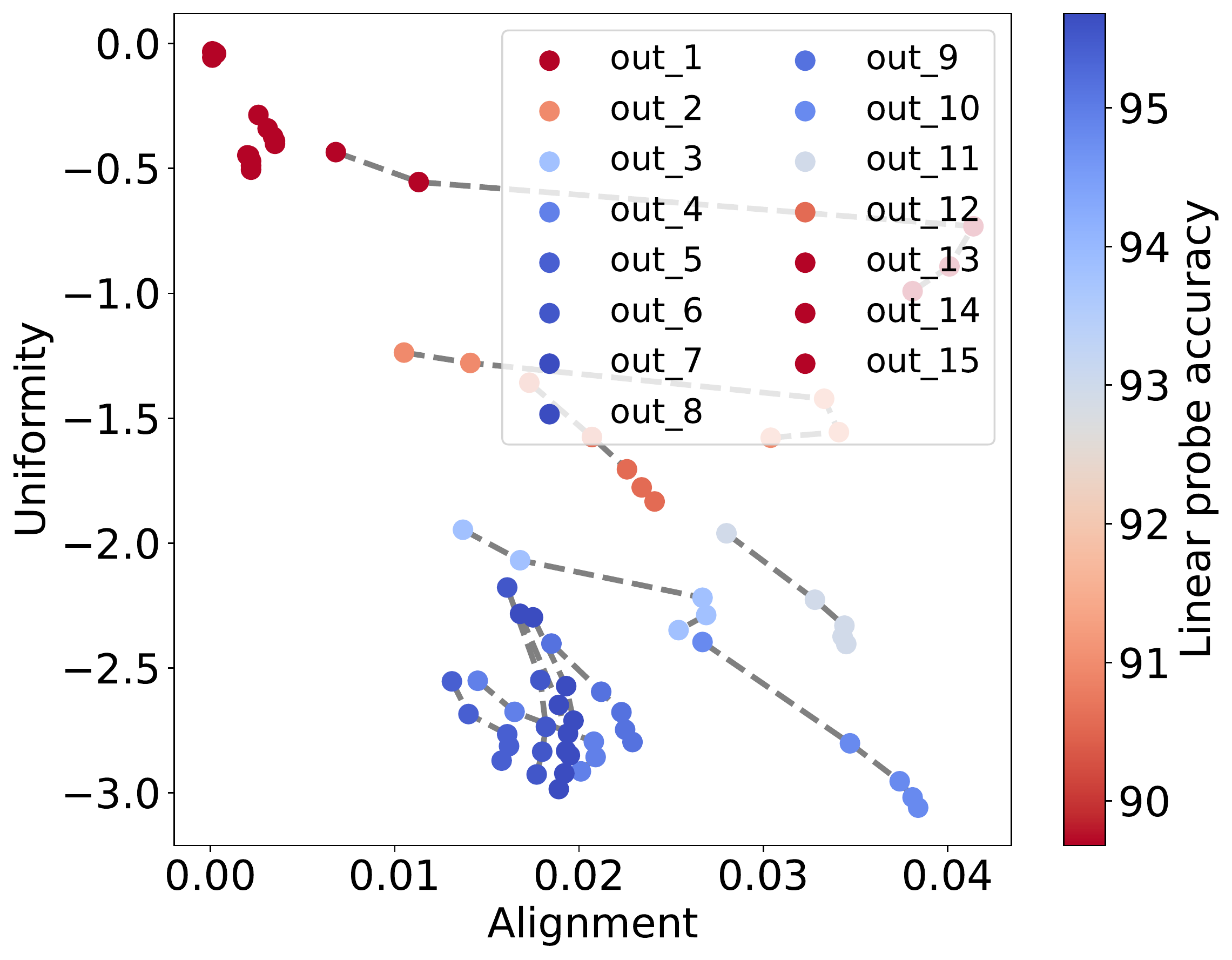}
    }
    \caption{\textbf{Alignment and uniformity metrics of features from different layers.}
    Dots and trend lines represent the training progress. The layers that produce strong linear probe results also obtain good and balanced alignment and uniformity, and the metrics are consistently getting optimized during the training progress.
    \textit{Left to right:} DDPM, DDPM++ trained by DDPM, and DDPM++ trained by EDM.
    }
\label{fig:align_uni}
\end{figure*}

Although DDAE learns high quality representations for discriminative tasks, it may rely on probing with annotated data to search for proper configurations. Therefore, it would be valuable if label-free metrics can indicate the best performing layers at training.
Inspired by feature distribution analysis \cite{alignuniform} in contrastive learning, we normalize the features from each layer to be on the unit hypersphere and investigate their alignment and uniformity \cite{alignuniform} properties.

In the context of contrastive learning, the alignment loss directly evaluates the distance between positive pairs, while the uniformity loss aims to measure the distribution uniformity on the unit hypersphere:
\begin{align*}
    \mathcal{L}_{\rm align}   &= \mathbb{E}_{(x,y) \sim p_{\rm pos}}{ [\|f(x) - f(y)\|_2^2] } \\
    \mathcal{L}_{\rm uniform} &= \log{ \mathbb{E}_{x,y \overset{\text{i.i.d.}}{\sim} p_{\rm data}}{ [e^{-2\|f(x) - f(y)\|_2^2}] } }
\end{align*}

To apply them to DDAEs, we need to define the positive pairs $(x,y) \sim p_{\rm pos}$ properly. Considering that the feature extraction method proposed in Section~\ref{sec:extracting} does not rely on deterministic noising, we assume that the representation should be independent to the noise sample $\epsilon$ used in Eq.~\ref{eq:noising}. To this end, we adapt these two metrics as:
\begin{align}
    \mathcal{L}_{\rm align}(t)   &= \mathbb{E}_{x \sim p_{\rm data},\epsilon_1,\epsilon_2}{ [\|f_t(x_t^{\epsilon_1}) - f_t(x_t^{\epsilon_2})\|_2^2] } \\
    \mathcal{L}_{\rm uniform}(t) &= \log{ \mathbb{E}_{x,y \overset{\text{i.i.d.}}{\sim} p_{\rm data},\epsilon}{ [e^{-2\|f_t(x_t^{\epsilon}) - f_t(y_t^{\epsilon})\|_2^2}] } }
\end{align}
where $f_t(\cdot)$ is the DDAE encoder, $x_t^\epsilon$ denotes the noised image using $\epsilon$, and $\epsilon_1, \epsilon_2$ are independently sampled noise.

We present the metrics with respect to selected layers in Figure~\ref{fig:align_uni}, to validate whether they align with the feature quality. We evaluate three DDAE implementations on the CIFAR-10 \textit{training} set: DDPM, DDPM++ trained by DDPM, and DDPM++ trained by EDM. For each model, $t$ is fixed to the respective optimal value, and we monitor the trajectories of metrics during training. Figure~\ref{fig:align_uni} shows that overall, $\mathcal{L}_{\rm align}$ and $\mathcal{L}_{\rm uniform}$ agree with the linear probe results (near lower left corners). Moreover, features from the well-performing layers show consistent improvements to both metrics during training. These results indicate that diffusion training may share similarities with contrastive representation learning, and selecting layers by metrics may mitigate the layer searching issue.

\begin{figure*}[t]
\centering
    \subfigure[Evaluation metrics of image generation and linear probing. Dots denote checkpoints at 150, 250, 500, 1000, 1500 and 2000 epochs. Images are sampled with the same seed.]{
    \includegraphics[width=0.55\linewidth]{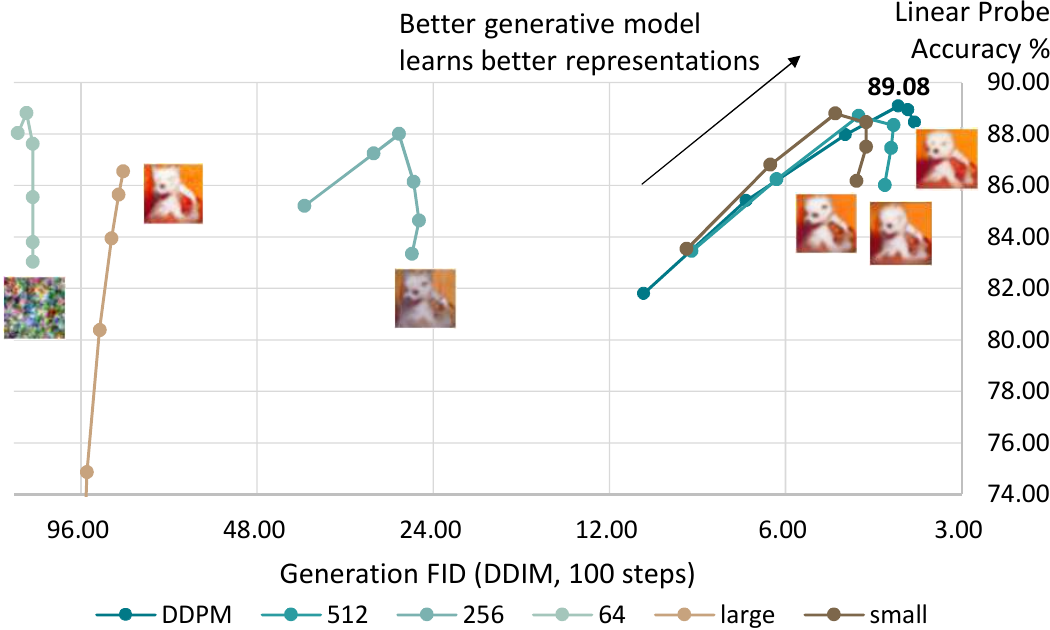}
    \label{fig:denoising_1}
    }
    \hfill
    \subfigure[Illustration of ablation configurations. Noise schedules are linearly spaced with varying numbers of levels and ranges.]{
    \includegraphics[width=0.42\linewidth]{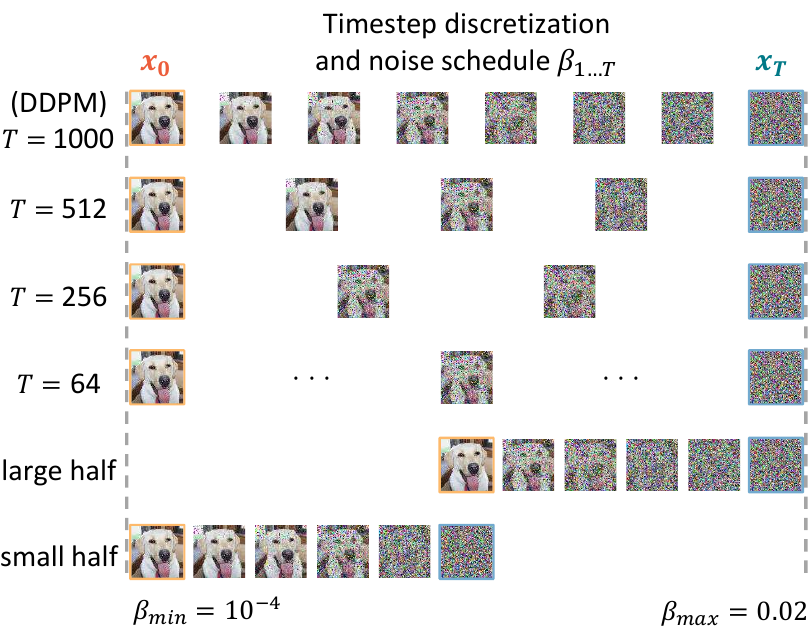}
    \label{fig:denoising_2}
    }
    \caption{\textbf{Correlation between generative and discriminative capabilities of DDAEs.}
    We ablate DDPM on CIFAR-10 from the perspective of denoising autoencoding. We observe that before over-fitting, better generative model learns better representations. Reduction on noise levels (shown in \textcolor[RGB]{1,121,135}{green}), noise ranges (shown in \textcolor[RGB]{125,103,75}{brown}) and training steps (denoted by dots) will weaken both performances.
    }
\label{fig:denoising}
\end{figure*}

\section{Experiments}
\label{sec:exp}
We firstly examine the impact of some core designs in diffusion models on both generative and discriminative performance through ablation studies. We then compare the results with counterparts on CIFAR-10 \cite{cifar10} and Tiny-ImageNet \cite{tiny} datasets under linear probing, fine-tuning and ImageNet \cite{in1k} transfer settings.

All models are retrieved or trained from official (or equivalent) codebases. For image classification, we do not use regularization methods such as mixup \cite{mixup} or cutmix \cite{cutmix}, and only employ lightweight augmentations. Implementation details as well as optimal layer-noise settings are provided in the code repository and the appendix.

\subsection{Main properties}
Previous research has demonstrated that in Transformers, better generative performance links to better representations, as measured by log-likelihood and linear probe accuracy \cite{igpt}.
Accordingly, we propose to investigate the correlation between generative and discriminative capabilities of DDAEs by plotting evaluation accuracy as a function of image quality calculated on 50k samples. In particular, the widely applied Fréchet Inception Distance (FID) \cite{fid} is used as the metric for generation quality.
We conduct ablation studies from two perspectives: (\romannumeral1) denoising autoencoding, and (\romannumeral2) generative pre-training, which correspond to the two sides of DDAEs as discussed in Section~\ref{sec:intro}.

\subsubsection{Denoising autoencoding}
Diffusion models can be viewed as multi-level denoising autoencoders. Based upon the findings in Figure~\ref{fig:layer_time} that DDAEs learn strongly linear-separable features in an unsupervised manner, we explore what design in the multi-level denoising makes DDAEs stronger representation learners than classical AEs, VAEs \cite{DRL-VDRL} and DAEs. In particular, we consider two key factors in DDPM \cite{ddpm} that may contribute to improved denoising pre-training: (1) the number of noise levels ($T$) and (2) the range of noise scales ($\beta_{1...T}$).

To investigate the effect of noise levels, we reduce the default $T=1000$ to $T=512$, $256$, and $64$. We opt not to decrease it further as DDPM cannot generate meaningful images when $T=64$. For each configuration, the noise schedule $\beta_{1...T}$ is linearly spaced in the range of $[\beta_{min}, \beta_{max}]$, where $\beta_{min}=10^{-4}$ and $\beta_{max}=0.02$, following DDPM default.
Additionally, we examine the effect of the noise scales by using the larger half and the smaller half of default $[\beta_{min}, \beta_{max}]$ schedule range. We keep $T=1000$ in these two scenarios for fair comparison.
Figure~\ref{fig:denoising_1} shows the influence of the noise configurations and training steps, and Figure~\ref{fig:denoising_2} provides an illustration of ablated noising configurations. DDPM with the maximum number of noise levels and the broadest noise scale coverage, achieves best performance in both image generation and recognition.

\begin{figure*}[t]
\centering
    \subfigure[DDAEs in DDPM and EDM trained on CIFAR-10.]{
    \includegraphics[width=0.48\linewidth]{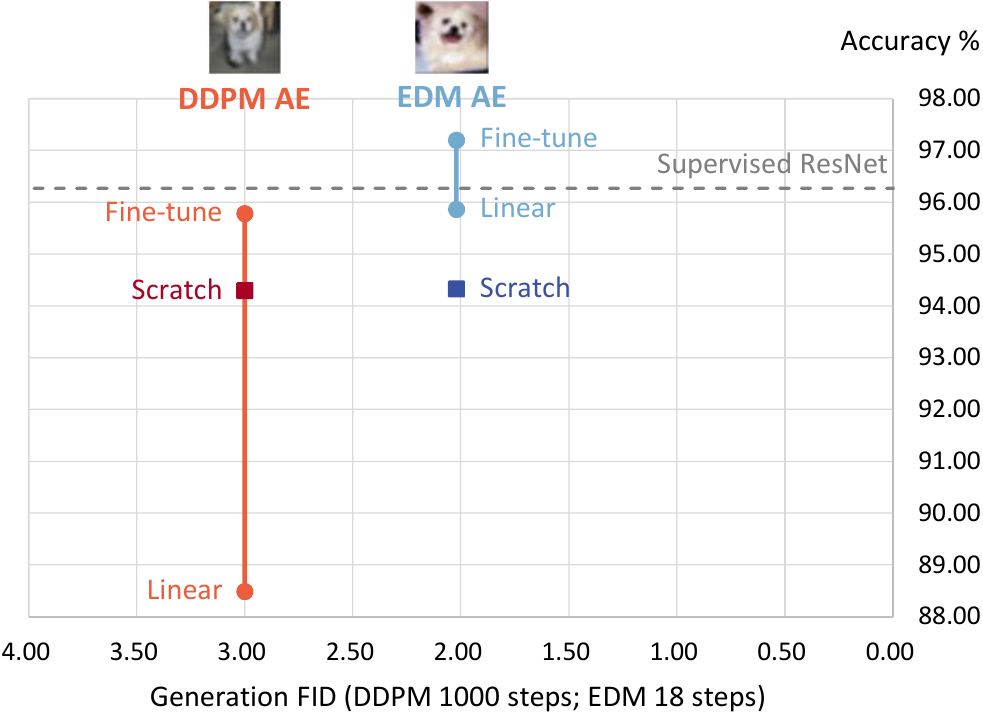}
    }
    \subfigure[DDAEs in DDPM and EDM trained on Tiny-ImageNet.]{
    \includegraphics[width=0.48\linewidth]{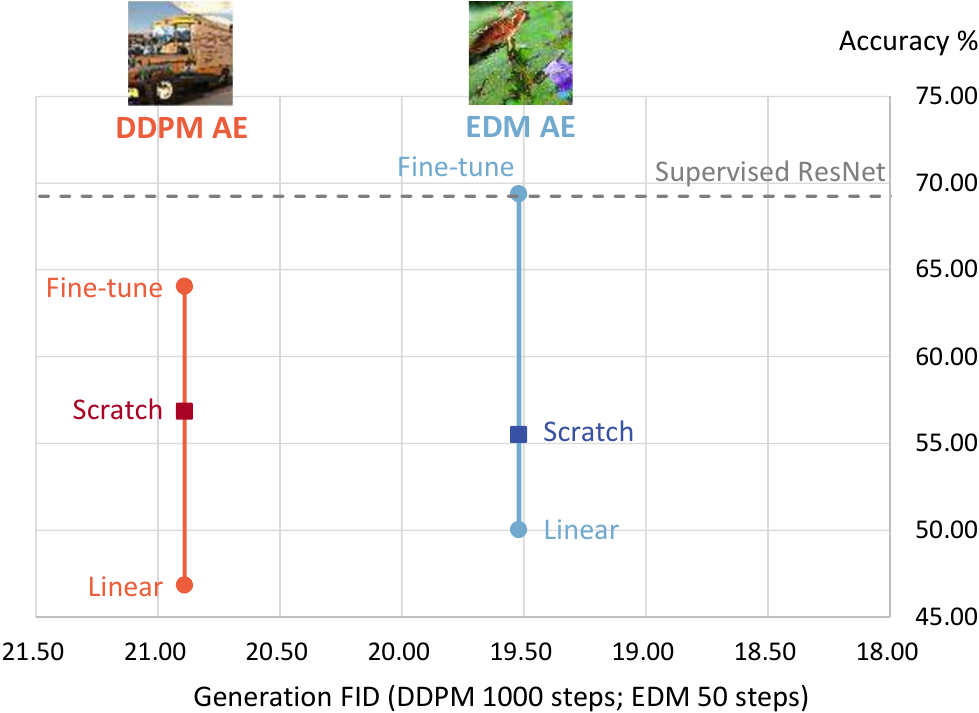}
    }
    \caption{\textbf{Generative and discriminative performance on multi-class datasets.} We evaluate DDAEs in two representative unconditional diffusion models. With improved training practice and network backbones, EDM-based DDAEs show better generative performance and promising linear probe accuracies. After fine-tuning, EDM-based encoders surpass supervised WideResNets.
    ``Scratch'' denotes the same truncated encoders trained from scratch, which are incompetent vision backbones and may have encumbered discriminative performance.
    }
\label{fig:generative-math}
\end{figure*}

\textbf{Noise level and range.} Reducing noise levels or narrowing the noise range weakens both generative and discriminative performances. Training handicapped diffusion models with only $T=256$ levels or the larger scales leads to more significant performance declines.
Interestingly, we observe that the recognition ability seems to depend less on a dense and wide noise configuration than generation. As shown in Figure~\ref{fig:denoising_1}, while generation FID may suffer huge decreases, the highest linear probe accuracy drops slightly by less than 3\%. The model with $T=64$ even produces stronger features than other handicapped models, despite its inability to generate meaningful images.

We attempt to explain these intriguing properties from a \textit{contrastive learning} angle.
Since the denoising objective (Section~\ref{sec:background}) demonstrates that DDAEs are forced to predict the very same $x_0$ given different versions of $x_t$ (Eq.~\ref{eq:noising}), there exists an underlying constraint of \textit{alignment} on these various noised versions. In other words, the time levels and random noises serve jointly as data augmentations, and different $x_t$s are positive views of $x_0$.
Consequently, a denser and wider noise configuration can increase the \textit{diversity of positive samples} and improve the representation quality. The high accuracy in the model with $T=64$ suggests that more noise levels could be an overkill for recognition-only purpose, but they still contribute significantly to generation.
Unfortunately, such diversity may require longer training duration --- Figure~\ref{fig:denoising_1} shows that models with half level counts or ranges can be well-trained at around 1000 epochs, while the full model requires 2000 epochs for this simple CIFAR-10. EDM \cite{edm} even relies on a 400 A100 GPU days training to reach state-of-the-art results on $64^2$ ImageNet. Due to the revealed duality of generation and recognition, we hope the best practices in discriminative representation learning (\eg self-distillation methods such as BYOL and DINO \cite{byol, dino}) will inspire future improvements to training efficiency in diffusion models beyond sampling, so that the scaling of diffusion models can be explored more efficiently. Introducing feature-level constraints to DDAEs could be helpful to accelerate training and boost discriminative performance, and we leave it to future work.

\textbf{Training step.} By tracking checkpoints throughout the training duration, it can be observed that better generative model learns better representations. Moreover, we observe that while linear probe accuracy tends to overfit after 1000 epochs in DDPM training, the generation performance continues to improve, indicating that it has not saturated. Similar observations can be found in other curves (Figure~\ref{fig:denoising_1}), that recognition tends to overfit earlier than generation.

Since diffusion training is a pixel-to-pixel task, it is reasonable to assume that DDAEs firstly learn high-level understandings from the noised input $x_t$ at the deeper layers, and gradually learn to predict the exact pixels through decoding. Consequently, models may focus on fitting some imperceptible details after learning saturated semantic representations. These properties somewhat provide support for the theory of relationship between image generation and understanding, as claimed intuitively in Section~\ref{sec:intro}.

In summary, by evaluating various checkpoints with different denoising pre-training setups, the positive correlation between the generative and discriminative capabilities of DDAEs is confirmed, aligning with the findings in iGPT.

\subsubsection{Generative pre-training}
To dive deeper into improved generative pre-training for recognition, we compare the basic 35.7M DDPM UNet with DDPM++ trained by EDM. With a larger network capacity, framework improvements, and data augmentation (CIFAR-10), we consider EDM as a better generative model, and examine whether it performs better in recognition similarly. Figure~\ref{fig:generative-math} shows the accuracies with respect to FIDs on two datasets.
Note that the networks and hyper-parameters for Tiny-ImageNet are far from optimal, since the best practice is too expensive, that it may cost 32 A100 GPU days to train a ADM network \cite{adm} on Tiny-ImageNet, according to EDM. Therefore, our goal is to provide confirmation of the observations, rather than optimizing performances.

\textbf{Recognition accuracy.} Figure~\ref{fig:generative-math} demonstrates that EDM outperforms DDPM on both generative and discriminative metrics.
On CIFAR-10, the linear probe accuracy increases dramatically from 88.5\% to 95.9\%. After fine-tuning, EDM achieves 97.2\% accuracy, surpassing WideResNet-28-10 \cite{wrn}. This superior classification performance confirms diffusion as a meaningful self-supervised pre-training approach.
On Tiny-ImageNet, observations are similar. While the FID has limited improvements between DDPM and EDM, the recognition rates increase more significantly by 3.2\% (linear probe) and 5.3\% (fine-tuning). The fine-tuned EDM also slightly surpasses supervised WideResNet.
The results suggest that improving diffusion models for better generation performance will naturally lead to better recognition models due to the effective generative pre-training.

\textbf{Network backbone issues.} Although the DDPM++ network in EDM has a larger model size than DDPM, their truncated versions fail to benefit from scaling when trained from scratch on both datasets (see ``Scratch'' in Figure~\ref{fig:generative-math}). Moreover, even though we tune hyper-parameters and train them longer, these truncated UNets fail to reach comparable accuracies as ResNets and are often unstable to train. These results suggest that truncating diffusion UNets at up-sampling and appending global pooling are not optimal practices for classification. These incompetent backbones for recognition may have encumbered the performance of DDAE. Using general-purpose vision backbones without up-sampling (\eg ViTs) or designing novel networks with explicit encoder-decoder split may overcome this issue.

However, ViT-based diffusion models are mainly operating in the latent space to achieve promising performance. We conduct preliminary experiments to compare pixel-space and latent-space image classification, and we find that the latter performs consistently worse on CIFAR-10 (96.3\% \textit{v.s.}~96.0\%) and Tiny-ImageNet (69.3\% \textit{v.s.}~65.3\%), suggesting that latent compression may lose information for recognition. Moreover, it may be an obstacle for downstream tasks such as object detection. We leave the exploration for unified pixel-space backbones to future work.

\begin{figure}[t]
\centering
    \subfigure{
    \includegraphics[width=0.9\linewidth]{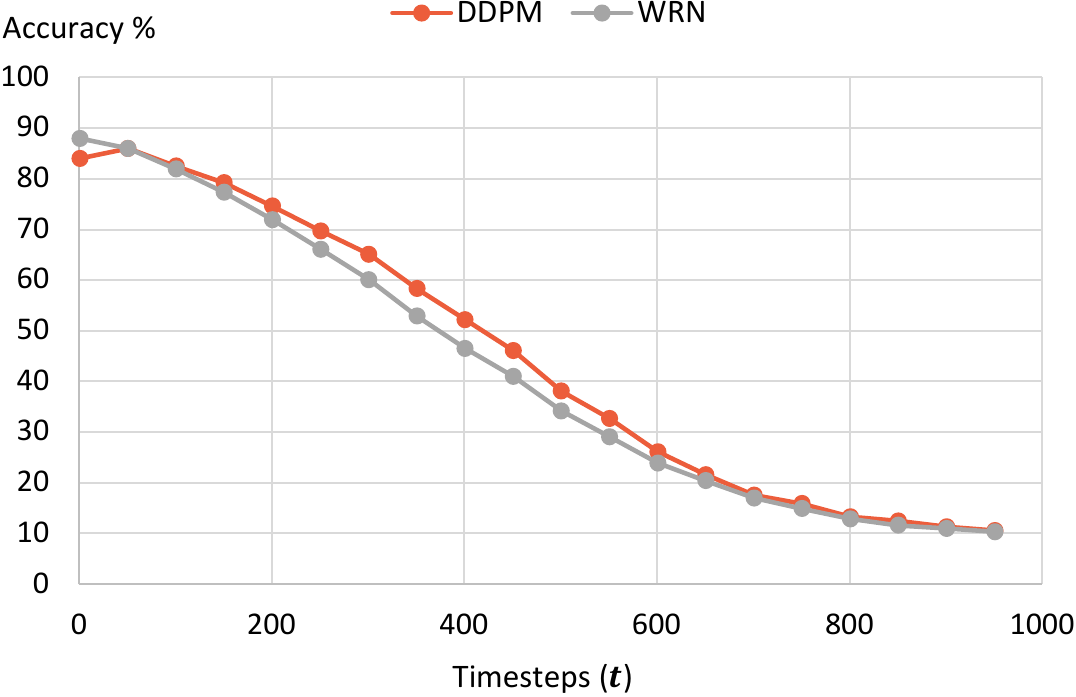}
    }
    \subfigure{
    \includegraphics[width=0.9\linewidth]{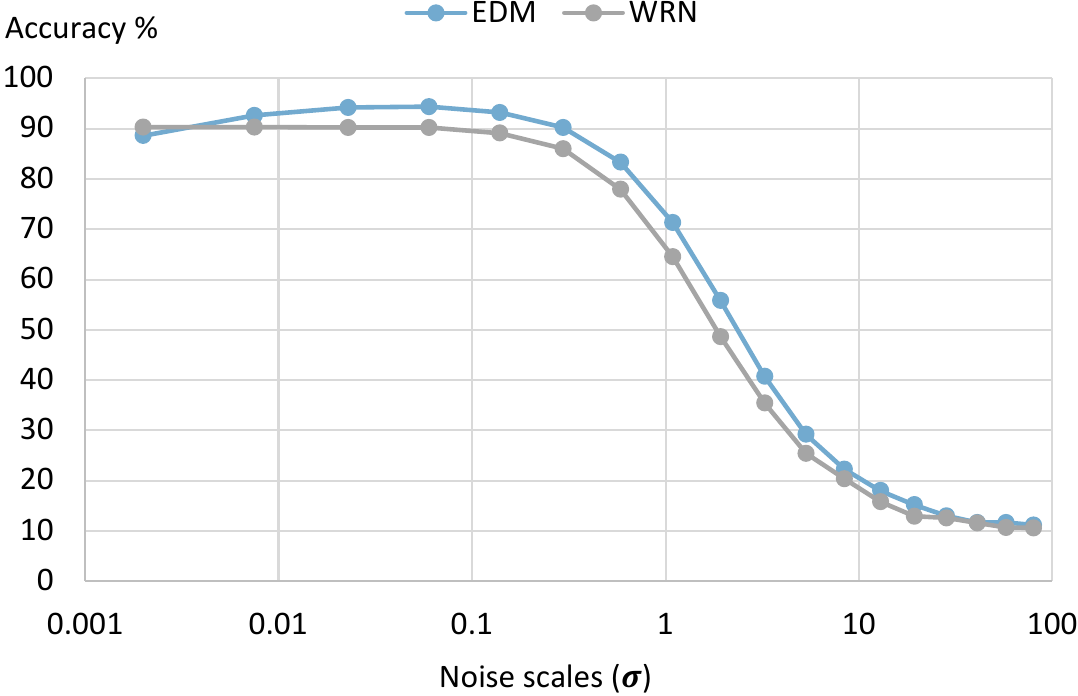}
    }
    \caption{\textbf{Accuracies of noise-conditional classifiers over different noise scales.}
    The two-layer classifier based on DDAE features achieves promising accuracies on noised image classification, and surpasses supervised models.
    \textit{Top:} DDPM and WideResNet (VP perturbation). \textit{Bottom:} EDM and WideResNet (VE perturbation).
    }
\label{fig:noisy}
\end{figure}

\subsubsection{Noise-conditional classifier}
Classifier guidance is a common manner to enhance conditional models \cite{sde, adm}, which relies on a noise-conditional classifier $\log{p_t(y|x_t)}$ to obtain the conditional gradients $\nabla_{x_t}{\log{p_t(x_t|y)}} = \nabla_{x_t}{\log{p_t(x_t)}} + \nabla_{x_t}{\log{p_t(y|x_t)}}$, or perturbed mean $\hat \mu(x_t|y) = \mu(x_t|y) + s\cdot \Sigma_t \nabla_{x_t}{\log{p_t(y|x_t)}}$.
Since the DDAE features are extracted with noise perturbation, our approach can naturally serve as such classifier. Specifically, we extract features in the same way as linear probing, and attach a two-layer MLP as the classifier head:
\begin{equation}
    p_t(y|x_t) = g(y|f_t(x_t) + \tau(t))
\end{equation}
where $f_t(\cdot)$ is the frozen DDAE encoder, $g(\cdot)$ is the two-layer classifier, and $\tau(t)$ denotes the timestep embedding.

Figure~\ref{fig:noisy} shows the accuracies of noise-conditional classifiers based on DDPM UNet and EDM-trained DDPM++. Following \cite{sde}, we train noise-conditional WideResNets on CIFAR-10 as supervised baselines and compare them to our DDAE-based approach. The curves show that a two-layer MLP head on frozen DDAE features can achieve promising accuracies, and outperforms supervised models over almost all noise scales. Most importantly, our simple approach can simultaneously obtain $\epsilon_\theta$ (or $D_\theta$, $\mu_\theta$) and the classification logits in \textit{one single} forward pass, reducing the time overhead in sampling caused by external classifiers.

\subsection{Comparison with previous methods}

\begin{table}[]
\resizebox{\linewidth}{!}{
    \begin{tabular}{llrr}
    \toprule
    \multirow{2}{*}{\textbf{Method}} & \multirow{2}{*}{\textbf{Evaluation}} & \multirow{2}{*}{\begin{tabular}[c]{@{}r@{}}\textbf{Generation}\\ \textbf{FID}\end{tabular}} & \multirow{2}{*}{\begin{tabular}[c]{@{}r@{}}\textbf{Acc.}\\ \textbf{\%}\end{tabular}} \\
     & & & \\ \toprule
    \multicolumn{4}{l}{\textit{on CIFAR-10}} \\
    \textcolor[RGB]{128,128,128}{WideResNet-28-10 \cite{wrn}} & \textcolor[RGB]{128,128,128}{Supervised} & \textcolor[RGB]{128,128,128}{N/A} & \textcolor[RGB]{128,128,128}{96.3} \\
    DRL/VDRL \cite{DRL-VDRL, DRL-VDRL-time} & Non-linear & $\sim$3.0 & \textless{}80.0 \\
    HybViT   \cite{hybrid}                  & Supervised & 26.4      & 95.9 \\
    SBGC     \cite{score-class}             & Supervised & *         & 95.0 \\
    \textbf{DDAE   (EDM)}                   & Linear     & 2.0       & 95.9 \\
    \textbf{DDAE   (EDM)}                   & Fine-tune  & N/A       & 97.2 \\ \midrule
    \multicolumn{4}{l}{\textit{on Tiny-ImageNet}} \\
    \textcolor[RGB]{128,128,128}{WideResNet-28-10 \cite{wrn}} & \textcolor[RGB]{128,128,128}{Supervised} & \textcolor[RGB]{128,128,128}{N/A} & \textcolor[RGB]{128,128,128}{69.3} \\
    HybViT   \cite{hybrid}                 & Supervised  & 74.8      & 56.7 \\
    \textbf{DDAE   (EDM)}                  & Linear      & 19.5      & 50.0 \\
    \textbf{DDAE   (EDM)}                  & Fine-tune   & N/A       & 69.4 \\                       
    \bottomrule
    \multicolumn{4}{l}{\begin{footnotesize}* Negative log-likelihood (NLL) of 3.11 is reported. Similar model \cite{sde} achieves \end{footnotesize}} \\
    \multicolumn{4}{l}{\begin{footnotesize} \enspace 2.99 NLL and 2.92 FID, for reference.\end{footnotesize}}
    \end{tabular}
}
\caption{\label{tab:diffusion-based-represent}\textbf{Comparisons with other diffusion-based representation learning methods on CIFAR-10 and Tiny-ImageNet.}
We compare DDAEs with unsupervised DRL/VDRL, supervised hybrid model HybViT, and supervised likelihood model SBGC. All results for other methods are retrieved from their original papers.
}
\end{table}

We compare EDM and DiT models with other diffusion-based representation learning methods and self-supervised discriminative methods. Since the generative performance is our first priority, we select checkpoints with the lowest FID for DDAEs, despite the recognition rates may overfit.

\textbf{Comparison with diffusion-based methods.} Table~\ref{tab:diffusion-based-represent} shows that EDM-based DDAE outperforms all previous supervised or unsupervised diffusion-based methods on both generation and recognition. Moreover, our DDAE can be seen as the state-of-the-art hybrid model \cite{hybrid} on CIFAR-10, which can generate and classify (through linear classifier) with a single model.
On Tiny-ImageNet, our self-supervised EDM yields significantly better generation FID than the supervised HybViT, despite a lower linear probe accuracy. After fine-tuning, DDAE catches up with supervised WideResNet and surpasses HybViT by large margins.

\textbf{Comparison with contrastive learning methods.}  Table~\ref{tab:cifar10} presents the evaluation results on CIFAR-10. For linear probing, EDM-based DDAE is comparable with SimCLRs considering model sizes. After fine-tuning, EDM achieves 97.2\% (w/o transfer) and 98.1\% (w/ transfer) accuracies, outperforming SimCLRs with comparable parameters, despite underperforming the scaled 375M SimCLR model by 0.5\%.
Table~\ref{tab:tiny} presents results on Tiny-ImageNet. Our EDM-based model significantly outperforms SimCLR pre-trained ResNet-18 under both linear probing and fine-tuning settings. However, DDAE is not as efficient as SimCLR on this dataset, that a slightly larger ResNet-50 can surpass our linear probe result with fewer parameters.

\textbf{Transfer learning with Vision Transformers.}
To verify the transfer ability on scalable ViTs, we transfer the DiT model, which is pre-trained on $256^2$ ImageNet, to CIFAR-10 and Tiny-ImageNet. Since the DiT codebase only provides the largest DiT-XL/2 checkpoint for class-conditional generation, we use it in an unconditional manner by dropping label to null \cite{cfg}. Although this may not be strictly fair due to the scale and supervision difference, we mainly aim to confirm the scalability of ViT-based DDAEs.

Table~\ref{tab:cifar10} and Table~\ref{tab:tiny} show that the scaled DiT-XL/2 outperforms the smaller MAE ViT-B/16 under all settings by large margins except for linear probing on CIFAR-10. It also catches up with the 375M SimCLR and achieves 98.4\% accuracy on CIFAR-10 after fine-tuning. These results indicate that similar to pixel-space ViTs, latent-space DiTs can also benefit from scaling and pre-training on larger datasets. However, diffusion pre-trained DiTs may not be as efficient as MAE pre-trained ViTs on recognition tasks, since the former is specifically designed for advanced image generation without optimizing its representation learning ability.

\begin{table}[]
\resizebox{\linewidth}{!}{
    \begin{tabular}{llrr}
    \toprule
    \multirow{2}{*}{\textbf{Method}} & \multirow{2}{*}{\textbf{Evaluation}} & \multirow{2}{*}{\begin{tabular}[c]{@{}r@{}}\textbf{Params}\\ \textbf{(M)}\end{tabular}} & \multirow{2}{*}{\begin{tabular}[c]{@{}r@{}}\textbf{Acc.}\\ \textbf{\%}\end{tabular}} \\
     & & & \\ \toprule
    \multicolumn{4}{l}{\textit{on CIFAR-10}} \\
    \textcolor[RGB]{128,128,128}{WideResNet-28-10 \cite{wrn}} & \textcolor[RGB]{128,128,128}{Supervised} & \textcolor[RGB]{128,128,128}{36} & \textcolor[RGB]{128,128,128}{96.3} \\
    \textbf{DDAE (EDM)}                 & Linear     & 36       & 95.9 \\
    SimCLR   Res-50 \cite{simclr}       & Linear     & 24       & 94.0 \\
    SimCLRv2 Res-101-SK \cite{simclr2}  & Linear     & 65       & 96.4 \\
    \textbf{DDAE (EDM)}                 & Fine-tune  & 36       & 97.2 \\
    SimCLRv2 Res-101-SK \cite{simclr2}  & Fine-tune  & 65       & 97.1 \\ \midrule
    \multicolumn{4}{l}{\textit{on CIFAR-10, with ImageNet transfer}}        \\
    \textbf{DDAE (EDM)}                 & Linear     & 36       & 91.4 \\
    SimCLR   Res-50 \cite{simclr}       & Linear     & 24       & 90.6 \\
    SimCLR   Res-50-4x \cite{simclr}    & Linear     & 375      & 95.3 \\
    \textbf{DDAE (EDM)}                 & Fine-tune  & 36       & 98.1 \\
    SimCLR   Res-50 \cite{simclr}       & Fine-tune  & 24       & 97.7 \\
    SimCLR   Res-50-4x \cite{simclr}    & Fine-tune  & 375      & 98.6 \\ \hdashline
    \textbf{DDAE (DiT-XL/2)\textsuperscript{\textdagger}} & Linear     & 314      & 84.3 \\
    MAE      ViT-B/16 \cite{MAE-report1}& Linear     & 86       & 85.2 \\
    \textbf{DDAE (DiT-XL/2)\textsuperscript{\textdagger}} & Fine-tune  & 314      & 98.4 \\
    MAE      ViT-B/16 \cite{MAE-report1}& Fine-tune  & 86       & 96.5 \\
    \bottomrule
    \multicolumn{4}{l}{\begin{footnotesize} \textsuperscript{\textdagger} Trained as class-conditional model but evaluated in an unconditional manner. \end{footnotesize}} \\
    \multicolumn{4}{l}{\begin{footnotesize} \enspace  Extra VAE encoder is used. \end{footnotesize}}
    \end{tabular}
}
\caption{\label{tab:cifar10} \textbf{Comparisons with other self-supervised methods on CIFAR-10.}
We compare DDAEs with contrastive SimCLRs, and masked autoencoders (MAE). Results for SimCLRs are retrieved from the original papers, and MAE's are reported by \cite{MAE-report1}. For DDAE and MAE, only encoder parameters are taken into account.
}
\end{table}

\begin{table}[]
\resizebox{\linewidth}{!}{
    \begin{tabular}{llrr}
    \toprule
    \multirow{2}{*}{\textbf{Method}} & \multirow{2}{*}{\textbf{Evaluation}} & \multirow{2}{*}{\begin{tabular}[c]{@{}r@{}}\textbf{Params}\\ \textbf{(M)}\end{tabular}} & \multirow{2}{*}{\begin{tabular}[c]{@{}r@{}}\textbf{Acc.}\\ \textbf{\%}\end{tabular}} \\
     & & & \\ \toprule
    \multicolumn{4}{l}{\textit{on Tiny-ImageNet}} \\
    \textcolor[RGB]{128,128,128}{WideResNet-28-10 \cite{wrn}} & \textcolor[RGB]{128,128,128}{Supervised} & \textcolor[RGB]{128,128,128}{36} & \textcolor[RGB]{128,128,128}{69.3} \\
    \textbf{DDAE   (EDM)}               & Linear     & 40       & 50.0 \\
    SimCLR   Res-18 \cite{report2}      & Linear     & 12       & 48.8 \\
    SimCLR   Res-50 \cite{report3}      & Linear     & 24       & 53.5 \\
    \textbf{DDAE   (EDM)}               & Fine-tune  & 40       & 69.4 \\
    SimCLR   Res-18 \cite{report3}      & Fine-tune  & 12       & 54.8 \\ \midrule
    \multicolumn{4}{l}{\textit{on Tiny-ImageNet, with ImageNet transfer}} \\
    \textbf{DDAE   (DiT-XL/2)\textsuperscript{\textdagger}} & Linear & 338 & 66.3 \\
    MAE   ViT-B/16 \cite{MAE-report1}   & Linear     & 86       & 55.2 \\
    \textbf{DDAE   (DiT-XL/2)\textsuperscript{\textdagger}} & Fine-tune & 338 & 77.8 \\
    MAE   ViT-B/16 \cite{MAE-report1}   & Fine-tune  & 86       & 76.5 \\
    \bottomrule
    \multicolumn{4}{l}{\begin{footnotesize} \textsuperscript{\textdagger} Trained as class-conditional model but evaluated in an unconditional manner. \end{footnotesize}} \\
    \multicolumn{4}{l}{\begin{footnotesize} \enspace  Extra VAE encoder is used. \end{footnotesize}}
    \end{tabular}
}
\caption{\label{tab:tiny} \textbf{Comparisons with other self-supervised methods on Tiny-ImageNet.}
We compare DDAEs with SimCLR and MAE. Results for SimCLR are reported by \cite{report2, report3} which are the highest in literature without cherry picking, and MAE's are from \cite{MAE-report1}.
}
\end{table}

\section{Discussion and conclusion}
We propose diffusion pre-training as a unified approach to simultaneously acquire superior generation ability and deep visual understandings, which potentially leads to the development of unified vision foundation models. However, as the first study to investigate diffusion for recognition at scale, there remain some limitations and open questions.

\textbf{Backbone issues.} Truncating DDAEs in the middle is not an elegant and optimal practice for encoders, and our approach relies on probing to find the best layer. Though the metric-based method works well on CIFAR-10, it may fail on more complex datasets where features are less linear-separable.
In constrast, ideal DDAE backbones may have explicit encoder-decoder disentanglement. Moreover, whether latent-based networks can rival pixel-space models on more recognition tasks needs more exploration.

\textbf{Efficiency issues.} Although DDAEs can achieve comparable accuracies to some pure recognition models, they rely on larger model sizes and are not efficient. Besides, diffusion models require longer training duration to achieve optimal generative performance, making them costly to scale.

\textbf{Relation to other self-supervised methods.} We hypothesize that the alignment between different images along the noising trajectory may implicitly contribute to the discriminative properties, which operates similarly to positive-only contrastive learning.
The concurrent study named Consistency Models \cite{CM}, which shares similarities to self-distillation-based contrastive learning \cite{byol, dino}, has already exploited this idea on denoising outputs. Moreover, we believe there exists another possibility of integrating such consistency or self-predictive constraints to DDAE features, resembling studies combining MAEs with contrastive methods \cite{cmae}.
It may also mitigate the previous issues by aggregating discriminative features on a designated encoder-decoder interface, and improving the learning efficiency.

\section*{Acknowledgements}
This work is partly supported by the National Key R\&D Program of China (2021ZD0110503), the National Natural Science Foundation of China (No. 62022011, 62202031, U20B2069), the Beijing Natural Science Foundation (No. 4222049), and the Fundamental Research Funds for the Central Universities.

{\small
\bibliographystyle{ieee_fullname}
\bibliography{egbib}
}

\clearpage
\appendix
\section{Implementation details}

\textbf{Diffusion pre-training.} We follow official implementations of DDPM, EDM and DiT for generative diffusion pre-training. The networks used in DDPM and EDM are UNets based on WideResNet with multiple convolutional down-sampling and up-sampling stages. Single head self-attention layers are used in the residual blocks at some resolutions. For CIFAR-10, we retrieve official checkpoints\footnote{\href{https://github.com/pesser/pytorch_diffusion}{https://github.com/pesser/pytorch\_diffusion}}\footnote{\href{https://github.com/NVlabs/edm}{https://github.com/NVlabs/edm}} from their codebases. For Tiny-ImageNet, we use official (or equivalent) implementations and similar configurations to train unconditional diffusion models by ourselves. The setting is in Table~\ref{tabapp:pretrain}.
Transformer-based DiT-XL/2 pre-trained on $256^2$ ImageNet is retrieved from its official codebase\footnote{\href{https://github.com/facebookresearch/DiT}{https://github.com/facebookresearch/DiT}}, and we do not train a smaller version (\eg DiT-B/2) due to the high computational cost. The used off-the-shelf VAE model for latent compression is retrieved from Stable Diffusion\footnote{\href{https://huggingface.co/docs/diffusers}{Hugging Face/Diffusers}}, which has a down-sample factor of 8.

\textbf{Linear probing and fine-tuning.} We use very simple settings for linear probing and fine-tuning experiments (see Table~\ref{tabapp:linear} and Table~\ref{tabapp:finetune}) and we intentionally do not tune the hyper-parameters such as Adam $\beta_1$/$\beta_2$ or weight decays.
In contrast with common practices in representation learning, we do not use additional normalization layers before linear classifiers since we find it also works well.

To train latent-space DiTs for recognition efficiently, we store the extracted latent codes through the VAE encoder and train DiTs in an offline manner. We encode 10 versions of the training set with data augmentations and randomly sample one version per epoch at the training. This approach may suffer from insufficient augmentation, and increasing augmentation versions or training with online VAE encoder may improve the recognition accuracy.

\textbf{Supervised training from scratch.} In Figure~\ref{fig:generative-math}, we present recognition accuracies of truncated UNet encoders trained from scratch and compare them to supervised Wide ResNets. The setting is in Table~\ref{tabapp:scratch}. We intentionally train these supervised models for long duration (200 epochs) to reach maximum performance for fair comparisons.

\section{Layer-noise combinations in grid search}
In Section~\ref{sec:extracting} we have shown that the layer-noise combination affects representation quality heavily. We perform grid searching to find a good enough, if not the best, combination for each model and dataset. For 18-step or 50-step EDM models, we train linear classifiers for 10 epochs with each layer and timestep. For 1000-step DDPM or DiT, we increase the timestep by 5 or 10 to search more efficiently. Table~\ref{tabapp:combi} shows the combinations adopted in Section~\ref{sec:exp}.

\begin{table}[]
\resizebox{\linewidth}{!}{
    \begin{tabular}{c|cc|cc}
    \hline
    \textbf{dataset}        & \multicolumn{2}{c|}{\textbf{CIFAR-10}} & \multicolumn{2}{c}{\textbf{Tiny-ImageNet}} \\
    \textbf{model}          & \textbf{DDPM}  & \textbf{EDM} & \textbf{DDPM}   & \textbf{EDM}    \\ \hline
    architecture            & DDPM           & DDPM++       & DDPM            & DDPM++          \\
    base   channels         & 128            & 128          & 128             & 128             \\
    channel   multipliers   & 1-2-2-2        & 2-2-2        & 1-2-2-2         & 1-2-2-2         \\
    attention   resolutions & \{16\}         & \{16\}       & \{16\}          & \{16\}          \\
    blocks   per resolution & 2              & 4            & 2               & 4               \\
    full   DDAE params      & 35.7M          & 55.7M        & 35.7M           & 61.8M           \\ \hline
    pre-training   epochs   & 2000           & 4000         & 2000            & 2000            \\ \hline
    \end{tabular}
}
\caption{\label{tabapp:pretrain} \textbf{Network specifications for diffusion pre-training.}}
\end{table}

\begin{table}[]
\resizebox{\linewidth}{!}{
    \begin{tabular}{l|lll}
    \hline
    \textbf{config}                    & \multicolumn{3}{l}{\textbf{value}}                              \\ \hline
    optimizer                          & \multicolumn{3}{l}{Adam with default momentum  \& weight decay} \\
    base   learning rate               & \multicolumn{3}{l}{1e-3}                                        \\
    learning   rate schedule           & \multicolumn{3}{l}{cosine   decay}                              \\
    batch size per GPU                 & \multicolumn{3}{l}{128}                                         \\
    GPUs                               & \multicolumn{3}{l}{4}                                           \\ \hline
    augmentations                      & \multicolumn{3}{l}{\texttt{RandomHorizontalFlip()} and}         \\
                                       & \multicolumn{3}{l}{\texttt{RandomCrop(32, 4)} for CIFAR-10 or}  \\
                                       & \multicolumn{3}{l}{\texttt{RandomCrop(64, 4)} for Tiny-ImageNet}\\ \hline
    \multirow[t]{4}{*}{training epochs}&                       & CIFAR-10         & Tiny-ImageNet        \\
                                       & DDPM                  & 10               & 20                   \\
                                       & EDM                   & 15               & 30                   \\
                                       & DiT                   & 30               & 30                   \\ \hline
    \end{tabular}
}
\caption{\label{tabapp:linear} \textbf{Linear probing setting.}}
\end{table}

\begin{table}[]
\resizebox{\linewidth}{!}{
    \begin{tabular}{l|lll}
    \hline
    \textbf{config}                    & \multicolumn{3}{l}{\textbf{value}}                              \\ \hline
    optimizer                          & \multicolumn{3}{l}{Adam with default momentum  \& weight decay} \\
    base   learning rate               & \multicolumn{3}{l}{1e-3 (DDPM and EDM), 8e-5(DiT)}              \\
    learning   rate schedule           & \multicolumn{3}{l}{cosine   decay}                              \\
    batch size per GPU                 & \multicolumn{3}{l}{128 (DDPM and EDM), 8 (DiT)}                 \\
    GPUs                               & \multicolumn{3}{l}{4 (DDPM and EDM), 8 (DiT)}    \\ \hline
    augmentations                      & \multicolumn{3}{l}{\texttt{RandomHorizontalFlip()} and}         \\
                                       & \multicolumn{3}{l}{\texttt{RandomCrop(32, 4)} for CIFAR-10 or}  \\
                                       & \multicolumn{3}{l}{\texttt{RandomCrop(64, 4)} for Tiny-ImageNet}\\ \hline
    \multirow[t]{3}{*}{training epochs}&                       & CIFAR-10         & Tiny-ImageNet        \\
                                       & DDPM                  & 30               & 80                   \\
                                       & EDM                   & 50               & 100                  \\
                                       & DiT                   & 50               & 50                   \\ \hline
    \end{tabular}
}
\caption{\label{tabapp:finetune} \textbf{Fine-tuning setting.}}
\end{table}

\begin{table}[]
\resizebox{\linewidth}{!}{
    \begin{tabular}{l|l}
    \hline
    \textbf{config}                    & \textbf{value}                              \\ \hline
    optimizer                          & Adam (DDAE encoder), SGD (WideResNet)      \\
    base   learning rate               & 5e-4 (DDAE encoder), 0.1 (WideResNet)      \\
    learning   rate schedule           & cosine   decay                              \\
    batch size per GPU                 & 128                                         \\
    GPUs                               & 4                                           \\ \hline
    augmentations                      & \texttt{RandomHorizontalFlip()} and         \\
                                       & \texttt{RandomCrop(32, 4)} for CIFAR-10 or  \\
                                       & \texttt{RandomCrop(64, 4)} for Tiny-ImageNet\\ \hline
    training epochs                    & 200                                         \\
    warmup epochs                      & 5                                           \\ \hline
    \end{tabular}
}
\caption{\label{tabapp:scratch} \textbf{Setting for training supervised models from scratch.}}
\end{table}

\begin{table}[]
\resizebox{\linewidth}{!}{
    \begin{tabular}{ll|ll}
    \hline
    \textbf{model} & \textbf{dataset@resolution} & \textbf{layer} & \textbf{timestep} \\ \hline
    DDPM  & CIFAR-10@32        & 7/12   (1st block@16)   & 11/1000  \\
    EDM   & CIFAR-10@32        & 6/15   (1st block@16)   & 4/18     \\
    DiT   & CIFAR-10@256       & 12/28                   & 121/1000 \\ \hline
    DDPM  & Tiny-ImageNet@64   & 2/12   (2nd block@8)    & 45/1000  \\
    EDM   & Tiny-ImageNet@64   & 7/20   (2nd   block@16) & 14/50    \\
    DiT   & Tiny-ImageNet@256  & 13/28                   & 91/1000  \\ \hline
    \end{tabular}
}
\caption{\label{tabapp:combi} \textbf{Adopted layer-noise combinations.} The numbers following ``@'' denote image or feature map resolutions.}
\end{table}

\end{document}